\documentclass[conference]{IEEEtran}
\IEEEoverridecommandlockouts                              

\usepackage{graphicx}
\usepackage[letterpaper, left=0.75in, right=0.75in, bottom=0.8in, top=0.75in]{geometry}

\usepackage{amsmath} 
\usepackage{amssymb}  
\usepackage{mathrsfs}
\usepackage{todonotes}
\usepackage{enumerate}
\usepackage{float}
\usepackage{caption}
\usepackage{subcaption}
\usepackage{cite}
\usepackage{tabularx}
\usepackage[ruled,vlined,linesnumbered]{algorithm2e}
\usepackage{bm}
\usepackage{xcolor}
\usepackage{mathrsfs}

\newcommand\R{\mathbb{R}}
\newtheorem{theorem}{Theorem}

\newtheorem{definition}{Definition}

\title{\LARGE \bf
Efficient Feature Mapping Using a Collaborative Team of AUVs
}

\author{\IEEEauthorblockN{Benjamin Biggs}
\IEEEauthorblockA{\textit{Virginia Tech}\\
babiggs@vt.edu}
\and
\IEEEauthorblockN{Daniel J. Stilwell}
\IEEEauthorblockA{\textit{Virginia Tech}\\
	stilwell@vt.edu}
\and
\IEEEauthorblockN{Harun Yetkin}
\IEEEauthorblockA{\textit{Virginia Tech}\\
yetkinh@vt.edu}
\and
\IEEEauthorblockN{James McMahon}
\IEEEauthorblockA{\textit{US Naval Research Laboratory}\\ Acoustics Division, Code 7130 \\
    james.mcmahon@nrl.navy.mil}

\thanks{*This work was supported by the Office of Naval Research via grants N00014-23-1-2345 and N00014-24-1-2267. The work of J. McMahon is supported by the Office of Naval Research through the NRL Base Program.}
\thanks{James McMahon is with the US Naval Research Laboratory, Code 7130, Washington D.C., USA}%
\thanks{Benjamin Biggs, Harun Yetkin, and Daniel Stilwell are with the Bradley Department of Electrical and Computer Engineering, Virginia Tech, Blacksburg, VA, USA, Harun Yetkin is also with Bartin University, Turkey}%

}

\begin{document}
\maketitle
\thispagestyle{empty}
\pagestyle{empty}

\begin{abstract}
 We present the results of experiments performed using a team of small autonomous underwater vehicles (AUVs)  to determine the location of an isobath. The primary contributions of this work are (1) the development of a novel objective function for level set estimation that utilizes a rigorous assessment of uncertainty, and (2) a description of the practical challenges and corresponding solutions needed to implement our approach in the field using a team of AUVs.  We combine path planning techniques and an approach to decentralization from prior work that yields theoretical performance guarantees.  Experimentation with a team of AUVs provides empirical evidence that the desirable performance guarantees can be preserved in practice even in the presence of limitations that commonly arise in underwater robotics, including slow and intermittent acoustic communications and limited computational resources.
 
\end{abstract}

\section{INTRODUCTION}

In the field of underwater robotics, autonomous underwater vehicles (AUV) can be employed for a variety of tasks such as mine hunting, collection of oceanographic data, and mapping of the ocean floor. The conventional approach to underwater feature mapping (e.g., mine-hunting) relies on an exhaustive search over an area with a single AUV. In this work, we seek to address the problem of deploying a team of AUVs that is able to collaboratively search an area of interest in a decentralized manner, allowing the area to be searched much more quickly and efficiently.

We specifically consider the problem of isobath localization or, in other words, the localization of a depth contour within a body of water. While our approach could be applied to other problems, isobath localization provides a simple way for us to demonstrate our work with a team of AUVs deployed in the field. A motivating application of isobath localization is determining the regions within a waterway that are safely navigable. The primary contributions of this work are (1) the development of a novel objective function for level set estimation that utilizes a rigorous assessment of uncertainty provided by Gaussian processes, and (2) a description of experiments with a team of AUVs that addresses the practical challenges of performing collaborative tasks with underwater robots.  In addition, we empirically evaluate if a team of AUVs deployed in the field obtains the performance guarantees that are due to the use of the multi-agent receding horizon path planning approach from \cite{biggs.etal.ICRA2021} and from recent results on the near-optimality of sequential (greedy) optimization of reward functions that are not necessarily submodular, such as ours, in \cite{biggs.etal.IROSS2022}.

Level set estimation is the process of using noisy observations of the function $f$ to determine the region(s) where $f$ exceeds some threshold \cite{WillettMinimaxLSE}. Level set estimation has applications in geospatial data \cite{sole2004morse}, bionformatics \cite{yang2002comparison}, and environmental studies \cite{szewczyk2004habitat, gotovos2013active, hitz2014fully}. This work is distinct from the existing literature on level set estimation in the following ways: (1) the level set is estimated using the decision-theoretic \textit{benefit of search} objective function presented in this work instead of other methods reported in the literature to measure classification ambiguity \cite{gotovos2013active, hitz2014fully, biggs.etal.ICRA2023}. We note that the work in \cite{biggs.etal.ICRA2023} is closely related to the work we report herein, but it only addresses the case of a single agent and uses a different reward function. 

For path planning, we formulate the isobath localization problem as an informative path planning (IPP) problem. Despite a host of techniques developed to provide approximate or suboptimal solutions to the informative path planning problem \cite{singh2009nonmyopic, schlotfeldt2018anytime, yetkiny2019online, kantaros2019asymptotically}, receding horizon methods are often used regardless of the planning techniques used within the shorter planning horizon \cite{atanasov2014information, atanasov2015decentralized, wakulicz2022informative}. However, the sequence of path segments in a receding horizon implementation do not inherit the near-optimality of each individual segment. The theoretical performance guarantees that we seek to leverage in this work rely on the use of a so-called terminal reward in the path planning optimization problem at each step in the construction of a receding horizon path. It is shown in \cite{biggs.etal.ICRA2021}  that the appropriate use of terminal rewards ensures that the reward attained by traversing sections of nearly-optimal paths (the complete receding horizon path) is no worse than a desired lower bound. 


\section{Search Objective for Level Set Estimation}\label{sec.lse}

The uncertain waterway depth $f$ is modeled by a Gaussian process, $f\sim \mathcal{GP}(m(p), k(p, p'))$.  An estimate of $f$ is computed from samples $d = (p,z)$ where $p \in \mathscr{P}$ is a location and  $z \in \mathbb{R}$ is a measuremet. Let $\mathcal{D}_t$ be the set of all samples acquired up to time $t$. For any input $p_i$, an associated measurement $z_i$ with
\begin{equation}
	z_i = f(p_i) + \epsilon_i, \ \text{for} \ i = 1, \ldots, |\mathcal{D}_t|
\end{equation}
where $|\cdot|$ denotes cardinality and $\epsilon_i \sim \mathcal{N}(0, \sigma_n^2)$ is drawn from a zero mean Gaussian distribution with standard deviation $\sigma_n$. 

Given the data set $\mathcal{D}_t$, one may wish to predict $f(p^*)$ for an arbitrary input $p^*$. From \cite[Chapter~2.7]{williams2006gaussian}, the posterior predictive distribution of $f(p^*)$ conditioned on $\mathcal{D}_t$ is Gaussian and specified by the mean $\mu_{p^*| \mathcal{D}_t}$ and variance $\sigma^2_{p^*| \mathcal{D}_t}$. 

We assume that incorrectly classifying a location as being above or below a threshold  incurs a loss. For example, falsely labeling a location as deep enough for ship navigation could lead to a collision while falsely labeling a location as too shallow for ship navigation could lead a vessel to travel further than necessary. Mathematically, we represent the costs incurred by poor estimation using a loss function $\mathcal{L}$.

In the context of isobath localization, $f(p)$ gives the true depth of the waterway at the input location $p$. For convenience, we use the notation $f_p \triangleq f(p)$.  Let $l \in \R$ be the level of interest. The indicator function
\begin{equation}\label{eq:indicator}
	g(f_p, l) = \begin{cases} 0 \ \text{if} \ f_p < l \\ 1 \ \text{if} \ f_p \geq l\end{cases}
\end{equation}
classifies the input location $p$ as being safe ($f_p \geq l$) or unsafe ($f_p < l$) for navigation. Let $\delta \in \{0, 1\}$ be an estimate of $g(f_p, l)$. We define our loss function for level set estimation as $\mathscr{L}(f_p, \delta) = c_i|g(f_p, l) - \delta|$ where $c_i > 0$ and
\begin{equation}\label{eq:depth_costs}
	c_i = \begin{cases} c_1 \ \text{if} \ g(f_p, l) < \delta \\
		c_2 \ \text{if} \ g(f_p, l) \geq \delta.
	\end{cases} 
\end{equation}
Intuitively, $c_1$ gives the cost associated with estimating a location to be safe when it is actually unsafe while $c_2$ gives the cost of estimating a location to be unsafe when it is actually safe.

\subsection{Bayes' Risk and the Benefit Of Search}\label{sec.benefit_of_search}
Recall that the true value of $f_p$ is not known. Therefore, the estimate $\delta$ which minimizes the loss function $\mathscr{L}(f_p, \delta)$ cannot be computed directly. However, because we assume $f$ is modeled by a Gaussian process, a probability distribution for $f_p$ is known. Let $\mathcal{D}_{t}$ be a data set and $\delta$ be an estimate of \eqref{eq:indicator}, the expected loss or \textit{risk} at an input point $p \in \mathscr{P}$ is
\begin{align}
	\mathbb{E}[\mathscr{L}(f_p, \delta) | \mathcal{D}_{t}] &= \int_{\R} \pi(f_p | \mathcal{D}_{t}) \mathscr{L}(f_p, \delta) \, df_p \\  \label{eq:lse_risk_1}
	&= \begin{cases}
		c_2\int_{l}^{\infty} \pi(f_p | \mathcal{D}_{t}) df_p  \ \text{if} \ \delta = 0 \\
		c_1\int_{-\infty}^l \pi(f_p | \mathcal{D}_{t}) df_p \ \text{if} \ \delta = 1.
	\end{cases}
\end{align}
where $\pi(f_p | \mathcal{D}_{t})$ is the Gaussian probability distribution of $f_{p}$ conditioned on the data set $\mathcal{D}_{t}$. Equation  \eqref{eq:lse_risk_1}  simply expands the notation and partitions the integral into the regions corresponding to the different regions where $f_p < l$ and $f_p \geq l$.

The Bayes' estimate is the estimate that minimizes the \textit{risk} in \eqref{eq:lse_risk_1} and is denoted
\begin{equation*}
	\delta^*(\mathcal{D}_t) = \underset{\delta \in \{0, 1\}}{\arg \min} \mathbb{E}[\mathscr{L}(f_p, \delta) | \mathcal{D}_{t}].
\end{equation*}
The \textit{Bayes' risk} is the expected loss given the Bayes' estimate
\begin{multline}
	\mathbb{E}[\mathscr{L}(f_p, \delta^*(\mathcal{D}_t)) | \mathcal{D}_{t}] = \\ \min \left( c_1\int_{-\infty}^l \pi(f_p | \mathcal{D}_{t}) df_p, c_2\int_{l}^{\infty} \pi(f_p | \mathcal{D}_{t}) df_p \right).
\end{multline}
To represent the Bayes' risk, we define
\begin{equation}
	r(\mathcal{D}_{t}) = \mathbb{E}[\mathscr{L}(f_p, \delta^*(\mathcal{D}_t)) | \mathcal{D}_{t}].
\end{equation}

\begin{definition}[Benefit of Search]\label{def.benefit_of_search}
	Let $\mathcal{D}_1$ and $\mathcal{D}_2$ be data sets. The \textit{benefit of search} is the reduction in Bayes' risk due to obtaining additional data and is written
	\begin{equation}\label{eq:benefit_of_search}
		\mathscr{B}(\mathcal{D}_2 | \mathcal{D}_1) = r(\mathcal{D}_1) - r(\mathcal{D}_1 \cup \mathcal{D}_2).
	\end{equation}
\end{definition}
Consider that a data set $\mathcal{D}_t$ is available at time $t$. Let $\mathcal{D}_{t+1:t+n}$ be the data set acquired in the $n$ time increments following time $t$. And let $\mathcal{D}_{t+n} = \mathcal{D}_{t} \cup \mathcal{D}_{t+1:t+n}$ be the combined data set. In planning at time $t$, the input locations $\mathcal{P}_{t+1:t+n}$ corresponding to the future data set $\mathcal{D}_{t+1:t+n}$ are known, but the measurements ($Z_{t+1:t+n}$) are not yet available. In planning, the benefit of search is computed using the expectation of the Bayes' risk
\begin{multline}\label{eq:expected_risk}
	\mathbb{E}_{Z_{t+1:t+n}}\left[r(\mathcal{D}_{t+n})\right] = \\ \int_{\R^{n_{\mathcal{V}}}}\int_{\R} \pi(f_p | \mathcal{D}_{t+n})\mathscr{L}(f_p, \delta^*(\mathcal{D}_{t+n})) df_p dZ_{t+1:t+n}
\end{multline}
where $n_{\mathcal{V}}$ denotes the cardinality of the measurement set $\mathcal{D}_{t+1:t+n}$ and, therefore, the dimensionality of the vector $Z_{t+1:t+n}$.

\begin{theorem}\label{thm.bofs_props}
	The benefit of search in \eqref{eq:benefit_of_search} is normalized and monotone in expectation.
\end{theorem}

Significantly, the normalized and monotone properties indicate that the benefit of search is an appropriate objective function for use with the greedy algorithm described in our prior work \cite{biggs.etal.ICRA2023} which provides a lower bound on greedily selected short-horizon paths when the reward function is normalized, monotone, but not submodular.   Due to space limitations, a proof of Theorem \ref{thm.bofs_props} is omitted here, but it can be found in \cite{biggs.dissertation.2023} Sec. 8.6.1. 

\subsection{Computing Expected Bayes' Risk}\label{sec.computing_risk}
While the expected Bayes' risk in \eqref{eq:expected_risk} may be approximated using Monte Carlo methods, a closed form approximation is desirable for real-time implementation. However, no closed-form solution to \eqref{eq:expected_risk} is known for this problem. Therefore, we propose an alternative approach. We address the computation of the expected Bayes' risk generally assuming a Gaussian process model. We note that measured values influence the posterior predictive mean but have no effect on the posterior predictive variance. Therefore, we consider a data set $\mathcal{Q}$ with $\mathcal{Q} = \mathcal{S} \cup \mathcal{V}$ for which the input locations $\mathcal{P}_{\mathcal{Q}}$ are known, but the measurement vector $Z_{\mathcal{V}}$ associated with the partial data set $\mathcal{V}$ is not yet available. 

We first determine the marginal probability of $\mu$ ($\pi(\mu_{p | \mathcal{Q}} | Z_{\mathcal{S}}, \mathcal{P}_{\mathcal{Q}})$).  Given the data set $\mathcal{Q}$, the posterior predictive distribution of $f_p$ is Gaussian with mean $\mu_{p | \mathcal{Q}}$, and variance $\sigma^2_{p | \mathcal{Q}}$. Marginalizing a product of Gaussian (see, e.g.,  \cite{biggs.dissertation.2023}, Sec. 8.71) yields the marginal probability of $\mu_{p | \mathcal{Q}}$ ($\pi(\mu_{p | \mathcal{Q}} | Z_{\mathcal{S}}, \mathcal{P}_{\mathcal{Q}})$) which is a Gaussian distribution with parameters
\begin{align}
	\mu_\mu &= \mu_{p | \mathcal{S}} \\
	\sigma^2_{\mu} &= \sigma^2_{p | S} - \sigma^2_{p | Q}.
\end{align}
Thus, the distribution for $\mu_{p | \mathcal{Q}}$ is centered around the posterior predictive mean given the set of available measurements with variance being the reduction in variance due to obtaining the data set $\mathcal{V}$ for which the measurement vector $Z_{\mathcal{V}}$ is not available.   The expectation of the Bayes' risk with respect to the marginal distribution of the posterior predictive mean can be expressed
\begin{equation} 
	\begin{split}
		&\mathbb{E}_{\mu}[r(\mathcal{Q})] \label{eq:expected_risk_mean} \\
		&= \int_{\R} \pi(\mu | Z_{\mathcal{S}}, 	\mathcal{P}_{\mathcal{Q}}) \int_{\R} \pi(f_p | \mu, \sigma^2_{p|\mathcal{Q}})\mathscr{L}(f_p, \delta^*(\mu)) df_p d \mu \\ 
		&= \int_{\R} \pi(\mu | Z_{\mathcal{S}}, \mathcal{P}_{\mathcal{Q}}) \\
		&\qquad \min 	\left( c_1 \pi(f_p < l | \mu, \sigma^2_{p|\mathcal{Q}}) , c_2 \pi(f_p \geq l | \mu, \sigma^2_{p|\mathcal{Q}}) \right)  d \mu
	\end{split}
\end{equation}
where $\delta^*(\mu)$ denotes the Bayes' estimate satisfying
\begin{equation}
	\delta^*(\mu) = \underset{\delta}{\arg \min} \ \int_{\R} \pi(f_p | \mu, \sigma^2_{p|\mathcal{Q}})\mathscr{L}(f_p, \delta) \, df_p.
\end{equation}

\noindent To handle the $\min$ function in \eqref{eq:expected_risk_mean}, we define the space
\begin{equation*}
	\Theta_{\mu} = \{\mu \in \R | c_1 \pi(f_p < l | \mu, \sigma^2_{p|\mathcal{Q}}) \leq c_2 \pi(f_p \geq l | \mu, \sigma^2_{p|\mathcal{Q}}) \}
\end{equation*}
or equivalently
\begin{equation*}
	\Theta_{\mu} = \left \{\mu \in \R  | \pi(f_p < l | \mu, \sigma^2_{p|\mathcal{Q}}) \leq \frac{c_2}{c_1 + c_2} \right \}.
\end{equation*}

\noindent Because $\pi(f_p < l | \mu, \sigma^2_{p|\mathcal{Q}})$ is simply the CDF of a Gaussian distribution, we have $\Theta_{\mu} = (-\infty, \mu^* ]$
where $\mu^*$ satisfies
\begin{equation}
	\pi(f_p < l | \mu^*, \sigma^2_{p|\mathcal{Q}}) = \frac{c_2}{c_1 + c_2}.
\end{equation}
Again due to space limitations, we refer to \cite{biggs.dissertation.2023}, Sec. 8.6.3 for the straight-forward steps that yields
\begin{equation}
	\mu^* = l - \text{erf}^{-1} \left(\frac{c_2 - c_1}{c_1 + c_2} \right)\sigma_{p|\mathcal{Q}} \sqrt{2}.
\end{equation}

\noindent The expected value of the Bayes’ risk conditioned on the posterior predictive may be expressed
\begin{multline}
	\mathbb{E}_{\mu}[r(\mathcal{Q})] = c_2\int_{-\infty}^{\mu^*} \pi(\mu | Z_{\mathcal{S}}, \mathcal{P}_{\mathcal{Q}}) \pi(f_p \geq l | \mu, \sigma^2_{p|\mathcal{Q}}) d \mu \\+ c_1\int_{\mu^*}^{\infty} \pi(\mu | Z_{\mathcal{S}}, \mathcal{P}_{\mathcal{Q}}) \pi(f_p < l | \mu, \sigma^2_{p|\mathcal{Q}}) d \mu
\end{multline}
Again saving space, we appeal to \cite{biggs.dissertation.2023}, Sec. 8.6.4, for the long but straightforward derivation of the expression,
\begin{multline}\label{eq:expected_risk_mostly_evaluated}
	\mathbb{E}_{\mu}[r(\mathcal{Q})] = \frac{(c_2 - c_1)}{4}\left[1 + \text{erf} \left( \frac{\mu^* - \mu_{\mu}}{\sigma_{\mu} \sqrt{2}} \right) \right] \\+ \frac{c_1}{2} + \frac{c_1}{2} \text{erf} \left( \frac{l - \mu_{\mu}}{\sqrt{2 \sigma^2_{p|\mathcal{Q}} + 2 \sigma_{\mu}^2}} \right) \\
	- \frac{(c_1 + c_2)}{2} \int_{-\infty}^{\mu^*} \pi(\mu | Z_{\mathcal{S}}, \mathcal{P}_{\mathcal{Q}})\text{erf} \left( \frac{l - \mu}{\sigma_{p|\mathcal{Q}} \sqrt{2}} \right) d\mu.
\end{multline}
To the best of the authors' knowledge, the solution to the integral in the final term of \eqref{eq:expected_risk_mostly_evaluated} is unknown and represents a topic of current research. We use the recently proposed \textit{exponential-quadratic approximation} \cite{pulford20224} to provide an approximate solution.  Again, due to limited space and due to the derivation of the approximation being straight-straightforward but long, we refer to \cite{biggs.dissertation.2023} Sec. 8.6.5 for details and the final form of the approximation.

\section{Path Planning}\label{sec.path_planning}
Given a team of $K$ agents, an $n$-length joint path $\gamma_n(s_i) = \{s_i, \ldots, s_{i+n} \}$ beginning at joint state $s_i$ is a feasible sequence of joint states. A joint state is a collection of individual agent states and is denoted $s_i = \{s_i^{1}, \ldots, s_i^{K} \}$. A sequence is feasible if there exists a joint action $a$ such that the team of agents will transition from $s_i$ to $s_{i+1}$ by executing the action $a$.

We assume that a state $s_i^k$ for agent $k$ corresponds to an input location $p_i^k \in \mathscr{P}$ and a joint state $s_i$ corresponds to a set of input locations $p_i = \{p_i^1, \ldots, p_i^K \}$. Let $\mathcal{P}_n(s_i)$ be the set of input locations corresponding to the $n$-length path $\gamma_n(s_i)$. We assume that, in traversing $\gamma_n(s_i)$, the team of agents obtains a vector of measurements $Z_n(s_i)$ corresponding to the input locations $\mathcal{P}_n(s_i)$ resulting in a data set $\mathcal{D}_{n}(s_i) = (\mathcal{P}_n(s_i), Z_n(s_i))$. 

Recall that the benefit of search is defined as a point-wise operation. Therefore, the reward associated with a path $\gamma_n(s_i)$ is a function of the locations at which the benefit of search is evaluated as well as the data set  $\mathcal{D}_{n}(s_i)$. Additionally, the reward associated with a path depends on previously obtained data $\mathcal{D}_{i-1}(s_0)$. 

We define the reward associated with a path $\gamma_n(s_i)$ to be the benefit of search at the locations along $\gamma_n(s_i)$ or
\begin{equation} \label{eq.optimal1}
	J(\gamma_n(s_i)) = \sum_{p_j^k}^{\mathcal{P}_n(s_i)} \mathscr{B}_j^k(\mathcal{D}_{n}(s_i) | \mathcal{D}_{i-1}(s_0))
\end{equation}
where $\mathscr{B}_j^k$ denotes the benefit of search defined in equation \eqref{eq:benefit_of_search} evaluated at the input location $p_j^k$. Our goal is for a group of agents to jointly maximize \eqref{eq.optimal1}


\subsection{Augmented Receding Horizon Path Planning}
We consider missions that correspond to path-length $l$ that is much too long to compute in real-time.  Thus we adopt a receding horizon approach where we compute $m$-length paths ($m<<l$) continuously.  A robot travels one step along a $m$-length path, and then computes another $m$-length path.  In \cite{biggs.etal.ICRA2021}, it is shown that the appropriate use of a terminal reward when computing $m$-length optimal paths yields lower bound on the value of the corresponding $l$-length receding horizon path. In this work, we assign the terminal reward to be the reward obtained by completing the mission (total length $l$) by naively mowing the lawn.  The lawn mower heuristic is  chosen because it is computationally efficient and is a common approach to subsea search and mapping applications.

 The remainder of this subsection describes the assumptions that must be made to directly obtain the lower bound guarantees in \cite{biggs.etal.ICRA2021}.  Let $\hat{\gamma}_{l-i-n-1}(s_{i+n+1})$ be a deterministic naive solution to the $l$-length path planning problem (e.g., mowing the lawn) beginning at the joint state $s_{i+n+1}$ which is feasible from the final state $s_{i+n}$ in $\gamma_n(s_i)$. Then we say that  $\bar{\gamma}_n(s_i) = \gamma_n(s_i) \cup \hat{\gamma}_{l-i-n-1}(s_{i+n+1})$ is the naively augmented path from the joint state $s_i$. We define the reward of a naively augmented path as
\begin{equation}\label{eq:augmented_path_value}
	\bar{J}_n(s_i) \triangleq  J(\gamma_n(s_i) \cup \hat{\gamma}_{l-i-n-1}(s_{i+n+1}))
\end{equation}

Consider the augmented path $\bar{\gamma}_n(s_i) = \{s_i, \ldots, s_{i+n}, \hat{s}_{i+n+1}, \ldots,  \hat{s}_l\}$. Using the property that for data sets $\mathcal{S}$ and $\mathcal{V}$, $\mathscr{B}(\mathcal{S} \cup \mathcal{V} | \emptyset) = \mathscr{B}( \mathcal{V} | \mathcal{S}) + \mathscr{B}( \mathcal{S} | \emptyset)$, the value of the $m$-length path $\gamma_m(s_i) \subset \bar{\gamma}_n(s_i)$ is
\begin{align}
	&J_m(s_i) = \bar{J}_n(s_i) -  \label{eq:aug_path_diff}\\
	& \sum_{p_j^k}^{\mathcal{P}_{l-m-1}(s_{i+m+1})} \mathscr{B}_j^k(\bar{\mathcal{D}}_{n-m-1}(s_{i+m+1}) \cup \mathcal{D}_{m}(s_i) | \mathcal{D}_{i-1}(s_0)) \notag \\ 
	& \quad - \sum_{p_j^k}^{\mathcal{P}_m(s_i)} \mathscr{B}_j^k(\bar{\mathcal{D}}_{n-m-1}(s_{i+m+1}) | \mathcal{D}_{i+m}(s_0)) \notag. 
\end{align}
The first term on the right-hand side of \eqref{eq:aug_path_diff} gives the reward for the augmented path $\bar{\gamma}_n(s_i)$. The second term gives the total benefit of search along the locations in $\bar{\gamma}_{n-m-1}(s_{i+m+1}) = \bar{\gamma}_n(s_i) \setminus  \gamma_m(s_i)$. The third term on the right-hand side of \eqref{eq:aug_path_diff} gives the additional reduction in risk at the locations in $\gamma_m(s_i)$ due to the data set acquired while traversing $\bar{\gamma}_{n-m-1}(s_{i+m+1})$.

Considering the third term on the right-hand side of \eqref{eq:aug_path_diff}, any additional reduction in risk due to the relatively small data set $\bar{\mathcal{D}}_{n-m-1}(s_{i+m+1})$ is likely negligible in practice. Therefore, we assume that 
\begin{align}\label{eq:augmented_benefit_at_future_path}
	&J_m(s_i) \approx \bar{J}_n(s_i) - \\
	& \sum_{p_j^k}^{\mathcal{P}_{l-m-1}(s_{i+m+1})} \mathscr{B}_j^k(\bar{\mathcal{D}}_{n-m-1}(s_{i+m+1}) \cup \mathcal{D}_{m}(s_i) | \mathcal{D}_{i-1}(s_0)) \notag.
\end{align}
Similarly, we assume that the second term on the right-hand side of \eqref{eq:augmented_benefit_at_future_path} may be appropriately approximated as
\begin{align}
	&\sum_{p_j^k}^{\mathcal{P}_{l-m-1}(s_{i+m+1})} \mathscr{B}_j^k(\bar{\mathcal{D}}_{n-m-1}(s_{i+m+1}) \cup \mathcal{D}_{m}(s_i) | \mathcal{D}_{i-1}(s_0)) \nonumber  \\ 
	&\approx \nonumber \\
        &\sum_{p_j^k}^{\mathcal{P}_{l-m-1}(s_{i+m+1})} \mathscr{B}_j^k(\bar{\mathcal{D}}_{n-m-1}(s_{i+m+1}) | \mathcal{D}_{m}(s_i) \cup \mathcal{D}_{i-1}(s_0))
\end{align}
meaning that the measurements obtained while traversing $\gamma_m(s_i)$ have little effect on the benefit obtained while traversing the locations along $\bar{\gamma}_{n-m-1}(s_{i+m+1})$. Given these assumptions, 
\begin{equation}\label{eq:aug_path_diff_2}
	J_m(s_i) \approx \bar{J}_n(s_i) - \bar{J}_{n-m-1}(s_{i+m+1})
\end{equation}
where the second term on the right-hand side of \eqref{eq:aug_path_diff_2} gives the reward for the augmented path $\bar{\gamma}_{n-m-1}(s_{i+m+1})$.

Given \eqref{eq:aug_path_diff_2}, Proposition 5.2 in \cite{biggs.etal.ICRA2021} guarantees that the value of a receding horizon path produced using augmented paths will be bounded below by $J(\hat{\gamma}_{l}(s_{0}))$ under the condition that the $n$-length augmented path $\bar{\gamma}_n(s_i)$ satisfies 
\begin{equation}\label{eq:bound_condition}
	\bar{J}_n(s_i) \geq \bar{J}_{n-m-1}(s_{i})
\end{equation}
at each planning step. 

The condition in \eqref{eq:bound_condition} is difficult to guarantee in practice, we discuss this briefly in Section  \ref{sec.conclusions}. However, it is shown in \cite{biggs.etal.ICRA2021} that a joint receding horizon path produced using the Dec-MCTS algorithm \cite{best2019dec} performs comparably to a method specifically designed to satisfy \eqref{eq:bound_condition} despite no explicit guarantees that \eqref{eq:bound_condition} is satisfied at each planning step. 
\section{Online Gaussian Process Regression}\label{sec.online_gpr}
Recall that the matrix inversion required for GP regression has time complexity $\mathcal{O}(|\mathcal{D}_t|^3)$ where $|\mathcal{D}_t|$ is the cardinality of the data set or the number of samples. Therefore, we seek a sparse representation of the environment with a limited number of kernel centers.  There are a number of well-known approaches to obtaining a sparse representation, including \cite{Csató_Opper_2002, Titsias, Nguyen-Tuong_Seeger_Peters_2009, Wilcox_Yip_2020}.  Because the focus of this work is on planning, and field implementation, we choose an effective and simple approach.  The maximum density of kernel centers, which can be related to the worst-case accuracy of prediction, is selected to limit the total number of kernel centers in a bounded area.  A datum $(p,z)$ is added to the set $\mathcal{D}_{t-1}$ only if $p$ is sufficiently far from all of the locations in $\mathcal{D}_{t-1}$.  Moreover, when computing a prediction at the location $p^*$, we use the data subset
$\hat{\mathcal{D}}_t = \{(p, z) \in \mathcal{D}_t : |p - p^*| \leq d_{\epsilon} \}$ where $d_{\epsilon}$ is related to the kernel length-scale.

\section{Field Trials}\label{sec.field_trials}
In order to demonstrate our augmented receding horizon path planning method in the field, we construct an example problem with the objective of localizing the 15 meter isobath within a bounded area of Claytor Lake in Dublin, VA, USA. 

\begin{figure}[H]
	\centering
	\includegraphics[width=\columnwidth]{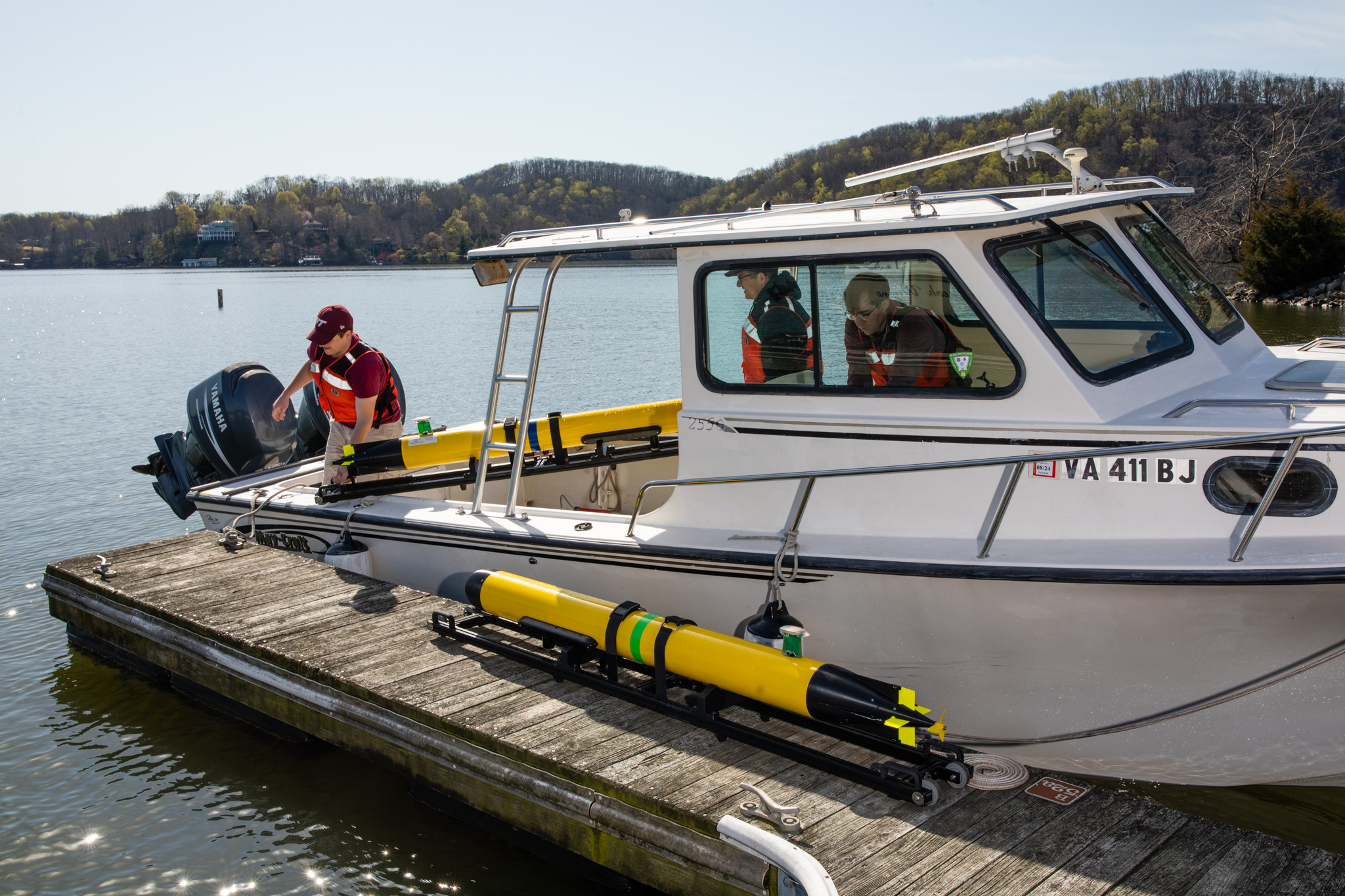}
	\caption{690 AUVs}
	\label{fig:690}
\end{figure}

In order to localize the isobath, teams of Virginia Tech 690 AUVs (shown in Figure \ref{fig:690}) were deployed and programmed to autonomously produce receding horizon paths that seek to maximize the path value function given in \eqref{eq:augmented_path_value} indirectly using sequential (greedy) optimization of the benefit of search reward function described in equation \eqref{eq:benefit_of_search}. Each 690 AUV is equipped with a suite of navigation sensors including a KVH 1750 inertial measurement unit (IMU), Teledyne Pathfinder Doppler velocity log (DVL), pressure sensor, WHOI micromodem 2 acoustic modem, and GPS receiver. The inertial navigation system (INS) on the vehicle uses an unscented Kalman filter on manifolds (UKF-M) \cite{brossard2020code} to fuse the data from each sensor into  an estimate of the vehicle’s attitude, velocity, and position as described in \cite{krauss.etal.OCEANS2022}. Additionally, each AUV is able to use its acoustic modem to communicate with other AUVs, although over severely limited bandwidth. 

Field trials were conducted using teams of 2 and 3 Virginia Tech 690 AUVs. For each team size, two receding horizon path planning methods are tested. One method utilizes a naive heuristic to augment short horizon paths while the other does not. The naive heuristic paths used in these field trials are simple lawnmower paths. All receding horizon paths produced are 100 steps long. The planning horizon $m$ for the method that includes naive heuristic paths is 3 steps. The planning horizon for the method that does not utilize augmented paths is 10 steps.  

The benefit of search gives the reduction in Bayes' risk due to obtaining a set of depth measurements at locations throughout the operational area. The depth of interest ($l$ in \eqref{eq:indicator}) is 15 meters. The costs of \eqref{eq:depth_costs} are equal in these experiments ($c_1 = c_2 = 10$). Therefore, the locations with the highest Bayes' risk throughout the operational area are associated with the expected location of the 15 meter isobath.

To prevent collision between AUVs, each AUV operates at its own fixed depth of 3, 4, and 5 meters for AUVs 1, 2, and 3 respectively.

In planning, the vehicle state is  $s = \{x_h, p_n, p_e \}$ where $x_h$ is the heading in radians and $p_n$ and $p_e$ are the North and East components of position. The set of next states available from any current state are based on a Dubins' type kinematic motion model. The onboard autopilot and guidance laws enable the AUVs to track kinematic paths.  The discrete dynamics defining the next state for an AUV are 
\begin{align}
	x_h[t+1] &= x_h[t] + a \\
	p_n[t+1] &=  \cos(-x_h[t])\Delta_x - \sin(-x_h[t])\Delta_y \\
	p_e[t+1] &= \sin(-x_h[t])\Delta_x + \cos(-x_h[t])\Delta_y.
\end{align}
where
\begin{align*}
	d & = r(\theta_{max} + |a|) \\
	\Delta_x &= \operatorname{sign}(a)[r - r \cos(|a|) + d \sin(|a|)] \\
	\Delta_y &= r \sin(|a|) + d \cos(|a|).
\end{align*}
for turning radius $r = 15$, a desired change in heading $
	a \in \mathcal{A} = \{-90, -30, -20, -10, 5, 0, 5, 10, 20, 30, 90 \}$,
and a maximum change in heading $\theta_{max} = 90$, where all angles are in degrees for convenience.


Joint path planning is accomplished via the sequential optimization (a greedy algorithm) wherein an ordering is assigned to the team of AUVs and each agent plans a path for itself considering only the paths of other AUVs that precede its position in the ordering (see, e.g., \cite{corah2017efficient}).  Additionally, because the benefit of search is both normalized and monotone, the greedy algorithm is expected to yield a high quality solution given the theoretical lower bounds presented in \cite{biggs.etal.IROSS2022} despite the benefit of search having no proven submodularity properties.

Each AUV plans paths for itself using the Monte-Carlo tree search method described in \cite{biggs.dissertation.2023} and \cite{biggs.etal.ICRA2021}  with each AUV seeking to maximize the additional reward that it contributes to the reward obtained by the paths of preceding AUVs per the sequential optimization (greedy) algorithm.

In order to facilitate communication between vehicles underwater using each vehicle's acoustic modem, a compact packet structure for communication of path and depth measurement data was designed. A time division multiple access (TDMA) scheme is used to prevent acoustic interference between vehicles while transmitting data. Each AUV is periodically allotted a 10 second window of time to transmit a single packet of data. The acoustic packets consist of 252 bytes. The first 14 bytes in a packet describe the initial state in the current best path known to the AUV. The initial state bytes are followed by a sequence of single bytes describing the actions from $\mathcal{A}$ taken to construct the path of the desired length. A terminal byte describes the end of the action sequence. When using naive paths to augment the short horizon paths, the terminal byte takes a value describing the deterministic heuristic used to produce the naive path. In this way, long paths may be communicated quite compactly. Following the terminal byte for the path, depth measurements are appended to the packet as groups of 12 bytes representing the north and east position of the measurement in the local tangent frame as well as the depth at that position. Given the packet size constraints, approximately 18 depth samples are communicated in a packet depending on the length of the short horizon path. To select which measurements to communicate, each AUV maintains a history of which measurements have been communicated and which have not. From the time-ordered list of samples that have not been communicated, the AUV selects every $n^{\text{th}}$ sample where $n$ is selected according to the number of samples that may be transmitted in a single packet.

\subsection{Results of Field Trials}\label{sec.results_of_field_trials}
\begin{figure}
	\centering
		\includegraphics[width=.85\columnwidth]{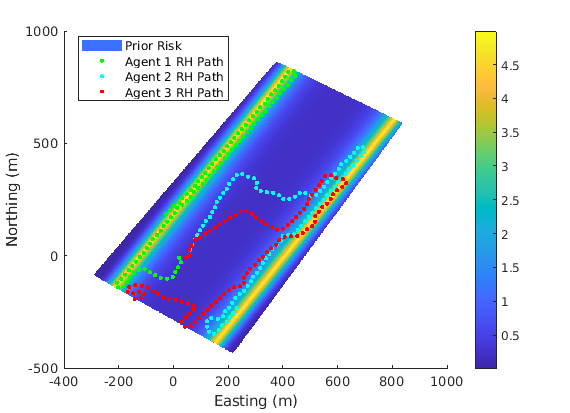} \\
		\includegraphics[width=.85\columnwidth]{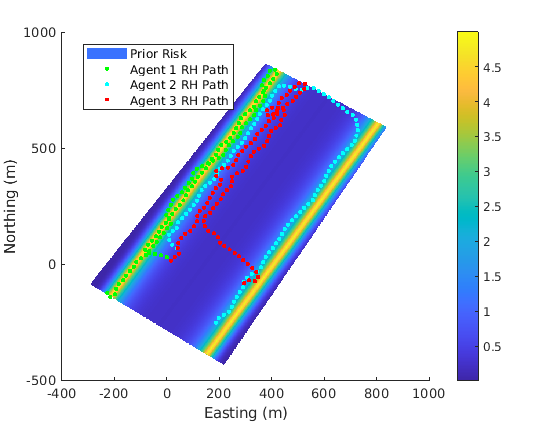} \\
		\includegraphics[width=.85\columnwidth]{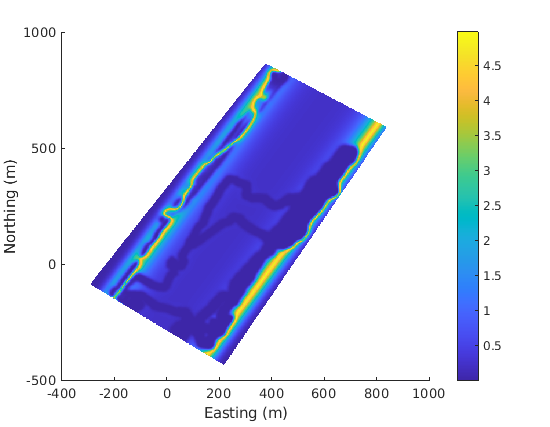} \\
		\includegraphics[width=.85\columnwidth]{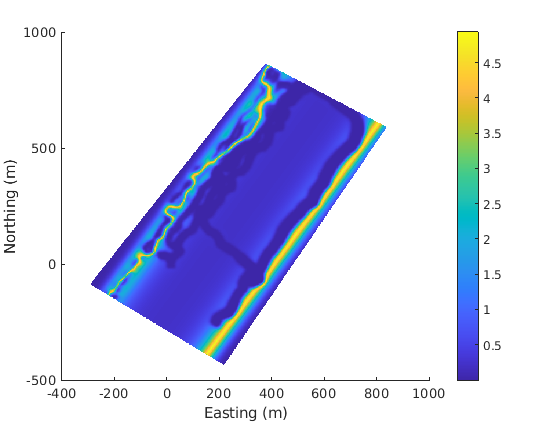} \\
	\caption{Paths and Bayes' risk for experiments accomplished using 3 vehicles.  Top two show the prior Bayes' risk while the bottom two panels show the posterior Bayes' risk.}
	\label{fig:3_agent_rh_paths}
\end{figure}

\begin{figure}
	\centering
	\includegraphics[width=\columnwidth]{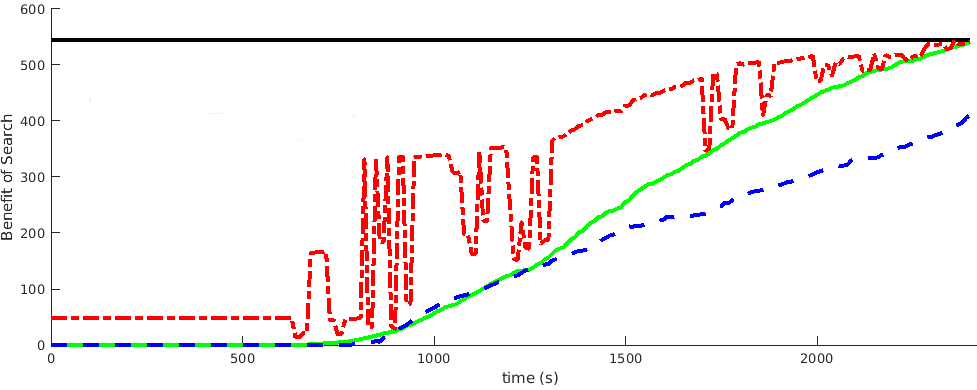}
	\caption{Results of receding horizon missions conducted with 3 vehicles; the green line is the accumulated reward with receding horizon planning and a terminal reward, the blue dash is the accumulated reward when no terminal reward is used, the black line is the guaranteed lower bound, which is the reward of a naive lawn-mower path, and the red dashed line is the cumulative reward of using receding-horizon planning with terminal reward plus the reward of the naive path.   }
	\label{fig:3_agent_results}
\end{figure}

The results for experiments conducted using 3 Virginia Tech 690 vehicles are presented in Figures  \ref{fig:3_agent_rh_paths} and \ref{fig:3_agent_results}.  In Figure \ref{fig:3_agent_rh_paths}, the prior Bayes' risk in the search environment for two experiments is shown in panels the top two panels. The bottom two panels show the corresponding posterior Bayes' risk given all measurements obtained by all AUVs while traversing the joint receding horizon path shown in the top two panels. For each path, all AUVs begin at the location (0,0).

Figure  \ref{fig:3_agent_results} shows the total reward accumulated by the team of AUVs for each planning method. The dashed blue line gives the total reward accumulated by the joint receding horizon path produced without the use of a naive path as a terminal rewards. The solid green line gives the total reward accumulated by the joint receding horizon path produced using a naive path as a terminal reward. The dashed red line gives the time-varying lower bound (the reward of the complete receding horizon path is bounded below by the reward of the naive path --- mowing the lawn) while the solid black line gives the maximum value of the time-varying lower bound. The receding horizon path produced using a naive path as a terminal reward performs no worse than the receding horizon path with no terminal rewards. In fact, it performs better. Given the non-idealities that arise when reducing our approach to practice, there is no guarantee that paths produced using the terminal rewards will perform better or worse than receding horizon paths produced using any other method.  However, the results of these experiments appear to demonstrate that the lower bound guarantee that the reward of the receding horizon path is bounded below by the reward of the naive path (mowing the lawn) may be achieved despite non-idealities such as those described in Section \ref{sec.conclusions}.

\section{Concluding Remarks}\label{sec.conclusions}

\begin{figure}
		\includegraphics[width=\columnwidth]{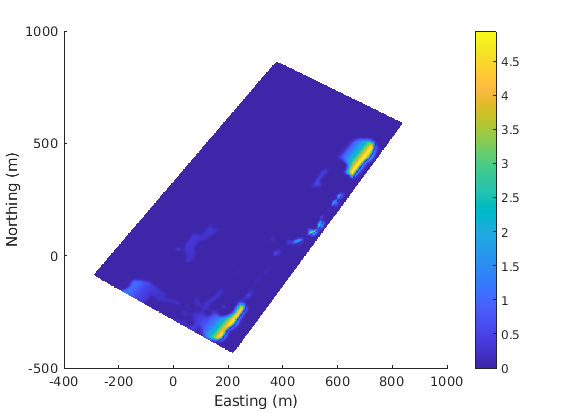}
		\label{fig:post_risk_error_1} \\
		\includegraphics[width=\columnwidth]{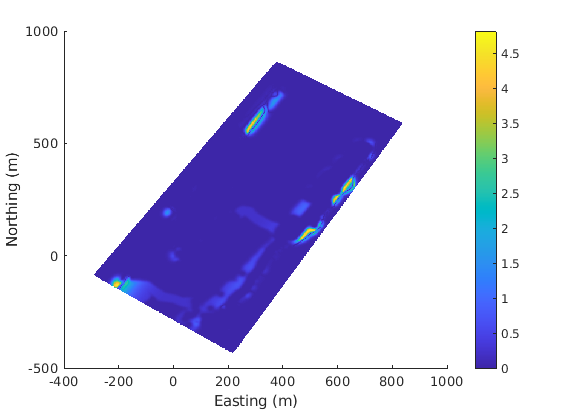}
		\label{fig:post_risk_error_2} \\
		\includegraphics[width=\columnwidth]{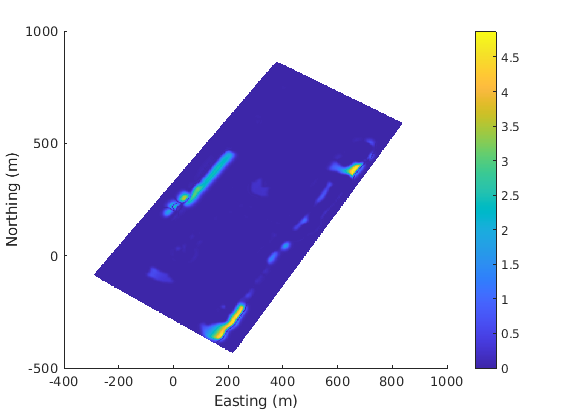}
 		\label{fig:post_risk_error_3} \\
	\caption{ The top, middle, and bottom panel show the difference between the posterior Bayes' risk given the measurements available to AUVs 1, 2, and 3, respectively, and the posterior Bayes' risk given all measurements shown in Figure \ref{fig:3_agent_rh_paths}. Note that all measurements not obtained by the AUV itself must be received via acoustic communications.}
	\label{fig:post_risk_error}
\end{figure}

The analysis in \cite{biggs.etal.ICRA2021} suggests that using a terminal reward when computing short-horizon segments of receding horizon paths, as in \eqref{eq:augmented_path_value}, should yield a total reward for the complete receding horizon path that is no less than would be obtained with no terminal reward.  Moreover, the results in  \cite{biggs.etal.ICRA2021} suggest that the complete path reward can be greater than when not using terminal rewards for case of computer simulations that are idealized, but do not fully meet the hypothesises in \cite{biggs.etal.ICRA2021}.  In this paper, we ask if the same property can be obtained by AUVs that are deployed in the field, in which case  additional practical challenges arise. 
 Indeed, a team of AUVs deployed in the field does not meet the requirements assumed in \cite{biggs.etal.ICRA2021} in several ways.
\begin{enumerate}

	\item \textbf{The greedy algorithm used in receding horizon path planning is not guaranteed to satisfy the condition of \eqref{eq:bound_condition}}: While the sequential optimization (greedy) algorithm as implemented in these field trials is not explicitly guaranteed to satisfy the condition of \eqref{eq:bound_condition}, the algorithm requires very little communication and is guaranteed to produce near optimal solutions when using an appropriate objective function as discussed in \cite{biggs.etal.IROS2020} and \cite{biggs.etal.ICRA2021}. 
	
	\item \textbf{Local approximations of the global objective function}: Each AUV only has access to a local approximation of the global objective function which accounts for all measurements obtained by all AUVs. Each local approximation accounts for all measurements obtained by the corresponding AUV, but the measurements obtained by all other AUVs are only available as they are communicated via acoustic channels. An example of the error between the posterior Bayes' risk for each AUVs vs the true posterior risk given all measurements from all AUVs is shown in Figure \ref{fig:post_risk_error}. Notably, given the limited communication bandwidth and the probability of data not being received at all, the error in each local approximation is small.
	
	\item \textbf{Slow communication prohibits ideal coordination in path planning}: The use of the TDMA protocol allows each AUV to transmit path planning data only once per TDMA cycle. Thus, a 10 second window ensures that each AUV is able to broadcast only once every 30 seconds given a team of 3 AUVs. In this time, an AUV may have an entirely different best path for planning.

 	\item \textbf{Asynchronous path traversal}: One AUV may traverse a short-horizon path faster than another; one AUV may reach the starting waypoint before another AUV and therefore begin the overall mission before the other AUV, etc.

\end{enumerate}
Fortunately, these disparities between theory and practice do not appear to have an appreciable negative effect on the results of these experiments with respect to the expected performance guarantees. 

\IEEEtriggeratref{10}
\bibliography{sources,stilwell}
\bibliographystyle{ieeetr}


\end{document}